\documentclass[letterpaper, 10pt, twocol]{ieeeconf}
\pdfoutput=1
\IEEEoverridecommandlockouts
\usepackage{amsmath, amssymb}
\usepackage[linesnumbered,ruled]{algorithm2e}
\usepackage{comment}
\usepackage[utf8]{inputenc} 
\usepackage[T1]{fontenc}    
\usepackage{textcomp}       
\usepackage{gensymb}        
\DeclareUnicodeCharacter{00B0}{\degree} 
\usepackage{graphicx}
\usepackage{hyperref}
\usepackage{subfig}
\usepackage{url}
\usepackage{fancyhdr}
\newcommand{\norm}[1]{\left\lVert#1\right\rVert}
\newcommand{\xx}{\mathbf{x}}
\newcommand{\uu}{\mathbf{u}}

\newcommand{\dxdt}{\frac{\D\xx(t)}{\D t}}
\newcommand{\D}{\mathrm{d}}

\title{Safe end-to-end imitation learning for model predictive control}
\IEEEpeerreviewmaketitle

\author{Keuntaek Lee\authorrefmark{1}\authorrefmark{2}, Kamil Saigol\authorrefmark{1}\authorrefmark{3} and Evangelos A. Theodorou\authorrefmark{2}
	\thanks{\authorrefmark{1}: Equal contribution. \authorrefmark{2}School of Electrical and Computer Engineering, the Institute for Robotics and Intelligent Machines, and the D. Guggenheim School of Aerospace Engineering, Georgia Institute of Technology, 270 Ferst Drive Atlanta GA 30332-0150, USA. \authorrefmark{3} Lyft Inc., San Francisco, CA, USA. Email: \tt\small keuntaek.lee@gatech.edu, \tt\small kamilsaigol@gatech.edu, \tt\small evangelos.theodorou@}
	\thanks{\tt\small gatech.edu}
}

\begin{document}

\maketitle
\thispagestyle{fancy}
\pagestyle{empty}

\begin{abstract}
We propose the use of Bayesian networks, which provide both a mean value and an uncertainty estimate as output, to enhance the safety of learned control policies under circumstances in which a test-time input differs significantly from the training set.  Our algorithm combines reinforcement learning and end-to-end imitation learning to simultaneously learn a control policy as well as a threshold over the predictive uncertainty of the learned model, with no hand-tuning required.  Corrective action, such as a return of control to the model predictive controller or human expert, is taken when the uncertainty threshold is exceeded.  We validate our method on fully-observable and vision-based partially-observable systems using cart-pole and autonomous driving simulations using deep convolutional Bayesian neural networks.  We demonstrate that our method is robust to uncertainty resulting from varying system dynamics as well as from partial state observability.
\end{abstract}

\section{Introduction} \label{sec:introduction}
As the deployment of deep neural networks as controllers for physical robotic systems becomes more prevalent, the issue of safety within artificial intelligence becomes an increasingly important concern.  Recently the use of end-to-end imitation learning to develop neural network control policies has surged in popularity, due in large part to the ease with which deep models can learn complex dynamics and infer global state from local data while bypassing the need for significant parameter tuning.  In contrast, traditional approaches to vision-based control rely on methods such image segmentation and object detection, classification, labeling, and filtering; often, these methods require significant engineering and tuning.  In addition, collecting the large datasets required by these methods is a difficult process, and it is often impossible to collect and process a dataset in a manner that is complete enough to account for every possible scenario.

The challenge, therefore, is to design a system capable of taking an appropriate action or making an appropriate prediction given novel data at test time, ideally with as little hand-tuning as possible. The ability to effectively handle novel data is key to introducing safety in intelligent control and is a particular weakness of current end-to-end methods of control.

It is easy to imagine situations in which novel data could be presented to a learned model at test time.  For example, malfunctioning sensors could pass unusual signals to a control system; changes in the environment could significantly affect system dynamics, such as in the case of robots that move from a smooth floor to a carpeted area; or, as is our motivation in this paper, a camera-based controller could encounter obstacles of a heretofore unseen type.

\begin{figure*}
\centering
\subfloat[Top: A sequence of input images for training. Bottom: A sequence of input images with novel obstacle at test time]{{\includegraphics[width=\textwidth]{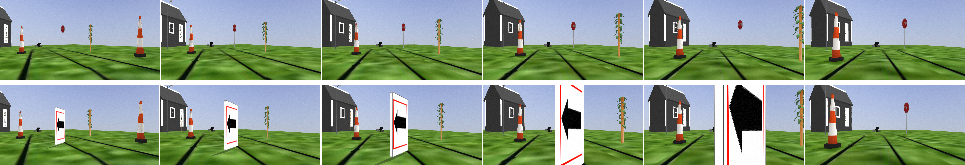}}}
\vspace{1mm}
\subfloat[Output mean and variance of control]{{\includegraphics[width=0.5\textwidth]{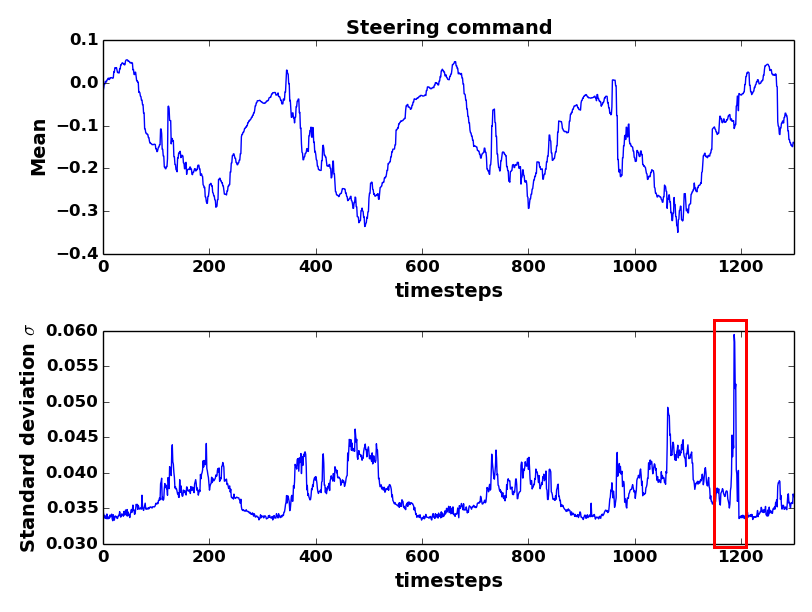}}}
\caption{Increased output variance from the Bayesian neural network due to a novel test-time input}
\label{fig:uncertainty increase}
\end{figure*}

Bayesian models -- that is, models that output predictive distributions, rather than point estimates -- have long been considered invaluable in machine learning for a number of reasons.  An estimate of model uncertainty can help to minimize risk, develop rejection criteria, compensate for extreme priors, and combine models \cite{Bishop2006}.  In the context of end-to-end learning, especially for autonomous vehicles and mobile robots, uncertainty estimates provide an easily interpreted safety measure.  By training models that can provide predictive uncertainty, we obtain a valuable indication of whether the model is able to produce a usable output; if it is not -- that is, if the uncertainty is high (Fig. \ref{fig:uncertainty increase}) -- we can take an appropriate action to compensate for the model's lack of knowledge.  In the context of end-to-end learning for autonomous driving such an action could include, for example, the return of control to a human driver.

Deep neural networks (DNN), popular due to their ability to model extremely complex functions, have enjoyed great success in a number of fields, including model-free reinforcement learning and model-based control.  However, while deep networks have been a revolutionary change to many engineering applications, their particular weakness is that they provide point estimates instead of a predictive distribution.  While undeniably effective, the use of DNNs results in a loss of the properties that make Bayesian methods so powerful.  As described in detail below, recent work on efficient deep Bayesian neural networks has helped to bridge this gap.

In this paper, we present an approach that combines deep end-to-end learning with Bayesian networks to learn high-speed model predictive control (MPC) while exploiting the predictive uncertainty of a Bayesian convolutional network to provide a measure of safety.  Our approach combines MPC with imitation learning and reinforcement learning (RL) methods to successfully perform complex tasks in both fully-observable and vision-based partially-observable scenarios while effectively handling novel data at test time.  Our approach is fully end-to-end; aside from the initial selection of network structure, comparatively little engineering or hand-tuning is required for success.

The rest of the paper is organized as follows: In Section \ref{sec:related_work}, we discuss work related to our method that appears in the literature.  Section \ref{sec:background} briefly introduces key background material.  In Section \ref{sec:safe_imlearn_control}, we define the problem of safe learning for MPC and present our solution.  Results of experiments are presented in Section \ref{sec:experiments}, followed by a discussion of future directions in Section \ref{sec:discussion}.  We conclude with Section \ref{sec:conclusion}.

\section{Related Work and Contributions} \label{sec:related_work}
Imitation and end-to-end learning are popular in the literature for a variety of tasks including the control of autonomous robots.  The DAgger meta-learning algorithm and its variants are popular due to the fact that these methods provide a regret bound that is linear in the time horizon \cite{DAgger}, contrary to earlier approaches with larger worst-case regret.  Earlier work, based on DAgger, specifically addressed the problem of learning a MPC policy for high-speed autonomous navigation \cite{Pan-NIPSWS-17}, but did not account for safety and robustness in the learned model.  In \cite{Pereira2018}, the authors develop an extension to the DAgger algorithm for end-to-end learning of MPC sequence policies with structured recurrent neural networks but do not address the problem of safe MPC.  In addition, partially-observable states and vision-based tasks are not considered.  In contrast, our work imposes no algorithmic structure on the neural network, avoids recurrent networks and the resulting difficulties in training, and successfully completes complex tasks without full state observability.  The NVIDIA PilotNet project \cite{Bojarski2017} does consider vision-based vehicle control with end-to-end learning and learns to recognize diverse obstacles and objects, but is not a probabilistic method and does not seem to explicitly handle truly novel test inputs.  Combining uncertainty measurement into the autonomous driving task, \cite{Strickland-17} investigated a predictive approach for collision risk assessment; however, this is a classification task rather than a regression task and cannot provide a learned controller for continuous action spaces.

The work most similar to our own is the DropoutDAgger approach \cite{DropoutDAgger}, a probabilistic extension to DAgger which uses the Dropout method of Bayesian approximation first presented in \cite{Gal2016Bayesian} to obtain a predictive distribution from a neural network.  The output variance is then used as a threshold to determine whether to return control to an expert controller, similar to the method presented herein.  However, in \cite{DropoutDAgger} both the dropout probability and the uncertainty threshold are selected manually; the tasks and experts considered are based on RL; and the case of partially observable environments and vision-based tasks is not considered at all.  In contrast, our work uses more principled methods to select dropout probabilities and the uncertainty threshold work.  In addition, we work in the MPC task domain rather than in the RL domain, which is more challenging due to the requirements for reactive, high-frequency control but is well-suited to deployment on real autonomous vehicles.  Finally, our method specifically addresses vision-based tasks using Bayesian convolutional neural networks.

Given the prior work in the area of end-to-end learning for control the contributions of this paper are summarized as follows:

\begin{enumerate}
\item We introduce the DropoutVGG Net, which combines VGG Net \cite{VGG}, the dropout technique for Bayesian approximation \cite{Gal2016Bayesian}, and Concrete Dropout \cite{Gal2017Concrete} for selection of the dropout probabilities in our neural network.
\item We use Cross-Entropy RL to optimally select an uncertainty threshold for return of control from a learned model to a MPC expert policy.
\item  Our formulation considers both the aleatoric and epistemic uncertainty of our neural network policy.
\item We demonstrate that our method is robust to uncertainty from varying system dynamics as well as uncertainty from partial state observability.
\item Use a model predictive controller as expert for a demonstration and successfully learn safe MPC using end-to-end learning with a Bayesian convolutional neural networks.  
\item We demonstrate our results on cart-pole balancing and autonomous driving tasks in both fully and partially observable cases.
\end{enumerate}

\section{Technical Background}
\label{sec:background}
In this section we briefly review key ideas used in our method.  Our formulation consists of an autonomous expert model predictive controller, a deep Bayesian neural network policy trained with imitation learning, and a reinforcement learning module to optimize the uncertainty threshold.

\subsection{Differential Dynamic Programming for Trajectory Optimization}
Iterative LQG/Differential Dynamic Programing (iLQG/DDP) \cite{MPCDDP} is a dynamic programming method which provides locally optimal solutions to the optimal control problem.  When used as the underlying optimizer for a model predictive controller, DDP provides real-time performance for complex, high-speed tasks; it also provides improved robustness relative to other methods due to the ability to calculate and return explicit trajectory-local feedback and feedforward gains.

Let the state of the system under consideration be $\xx \in \mathbb{R}^n$ with controls $\uu \in \mathcal{U} \subset \mathbb{R}^m$.  We then consider the following optimal control problem.

\begin{align}\label{eq:optcontrolprob}
	&V(\xx(t_0), t_0) = \min_{\uu(t)}\Big[\phi(\xx(t_f), t_f) + \int_{t_0}^{t_f}l(\xx(t), \uu(t), t)\D t\Big] \\
	&\mbox{subject to dynamics}~ \dxdt = f(\xx(t), \uu(t), t)
\end{align}

Here $V$ is the value function, the time horizon is $[t_0, t_f]$, and we have terminal cost $\phi$ with running cost $l$.  Let nominal state and control sequences be given by $\bar{\xx}({\cdot}$) and $\bar{\uu}(\cdot)$, respectively, so that we can define the state and control sequence updates $\delta\xx$ and $\delta\uu$ such that $	\xx = \bar{\xx} + \delta\xx $ and $  	\uu = \bar{\uu} + \delta\uu $. 
DDP provides an efficient iterative method to compute the trajectory updates.  To begin, note that discretization and linearization of the dynamics about the nominal trajectory results in the discrete, time-varying linear system

\begin{align}\label{eq:linear_expansion}
\delta\xx(t_{k+1})
&= \Phi\delta\xx(t_k) + B\delta\uu(t_k),
\end{align}

where we have defined $\Phi = \Big[I + \frac{\partial f}{\partial \xx}(\bar{\xx}(t_k), \bar{\uu}(t_k), t_k)\Delta t\Big]$ and $B = \frac{\partial f}{\partial \uu}(\bar{\xx}(t_k), \bar{\uu}(t), t_k)\Delta t$. Applying the Bellman principle to \ref{eq:optcontrolprob} and discretizing, we obtain

\begin{align}\label{eq:discrete_bellman}
	V(\xx(t_k), t_k) = \min_{\uu(t_k)} \Big [\underbrace{l(\xx(t_k), \uu(t_k))\Delta t + V(\xx(t_{k+1}), t_{k+1})}_{Q} \Big ]. 
\end{align}

The prime notation on $V$ and derivatives indicates evaluation at $t_{k+1}$.  Taking a quadratic expansion of the right hand side of \ref{eq:discrete_bellman} and then noting that $\uu$ and $\delta\uu$ have a linear relationship so that we can perform the minimization in \ref{eq:discrete_bellman} over $\delta\uu$, we obtain the optimal control update law

\begin{equation}\label{eq:opt_control_update}
	\delta\uu(t_k) = -Q_{uu}^{-1}(Q_u + Q_{ux}\delta\xx(t_k)),
\end{equation}

where the first and second derivatives of $Q$ are

\begin{align}\label{eq:Q_and_derivs}\begin{split}
Q_0 &= l\Delta t + V', Q_x = l_x\Delta t + \Phi^T V'_x,~	Q_u = l_u \Delta t + B^T V'_x \\
Q_{xu} &= l_{xu}\Delta t + \Phi^TV'_{xx}B,~	Q_{ux} = Q_{xu}^T \\
Q_{xx} &= l_{xx} \Delta t + \Phi^TV'_{xx} \Phi,~ Q_{uu} = l_{uu} \Delta t + B^TV'_{xx}B.
\end{split}
\end{align}

It is important to note that the optimal control update is given by a feedback control law with feedback gain $K_{fb} = -Q_{uu}^{-1}Q_{ux}$ and feedforward gain $K_{ff} = -Q_{uu}^{-1}Q_u$.

For each iteration of DDP, updates along a trajectory are computed using the dynamic programming methodology, starting at the terminal time and computing optimal updates backward along the trajectory.  The equations for this backward pass are computed by substituting \ref{eq:opt_control_update} into the quadratic expansion of \ref{eq:discrete_bellman} and grouping terms of corresponding order to obtain

\begin{align}
V(\bar{\xx}(t_k), t_k) &= Q_0 - \frac{1}{2} Q_u^T Q_{uu}^{-1} Q_u \label{eq:bw_first}\\
V_x(\bar{\xx}(t_k), t_k) &= Q_x - Q_{xu} Q_{uu}^{-1} Q_u \\
V_{xx}(\bar{\xx}(t_k), t_k) &= Q_{xx} - Q_{xu} Q_{uu}^{-1} Q_{ux}. \label{eq:bw_last}
\end{align}

After the backward pass is complete, a forward pass finds the new nominal control and state trajectories using the computed gains and the nonlinear dynamics, respectively.  The backward and forward passes are iterated until the maximum number of iterations is exceeded or until convergence.  DDP is summarized in Algorithm \ref{alg:algorithm_ddp} .  In our work, we apply DDP as part of the model predictive control algorithm by performing a small number of DDP iterations at each time step, applying the first control of the resulting optimal sequence to the plant, obtaining a new state estimate, and repeating until the task is complete.

\begin{algorithm}
	\caption{Differential Dynamic Programming}\label{alg:algorithm_ddp}
	\SetKwInOut{Input}{Input}
	\SetKwInOut{Output}{Output}
	
	\Input{$\bar{\xx}_0$ : initial state, $\bar{\uu}_{0:H-2}$ : control sequence, \\
		$H$ : time horizon, $\Delta t$ : time step size, \\
		$i_{max}$ : max iterations, $\alpha$ : learning rate \\
		$f$ : transition function, \\
		$l, \phi$ : running and terminal cost functions} 
	\Output{$\bar{\uu}_{0:H-2}$ : optimal control sequence, \\
			$\bar{\xx}_{0:H-1}$ : optimal state trajectory,\\
			$K^{fb}_{0:H-2}, K^{ff}_{0:H-2}$ : control gains
		}
	
	\For{$t \gets 1$ \KwTo $H-1$}
	{
		$\bar{\xx}_t \gets \bar{\xx}_{t-1} + f(\bar{\xx}_{t-1}, \bar{\uu}_{t-1}) \Delta t$ \\
		Calculate $\Phi$, $B$ at $t$ using \ref{eq:linear_expansion}
	}
	
	\For{$i \gets 0$ \KwTo $i_{max}-1$}
	{
		$V_{H-1} \gets \phi(\bar{\xx}_{H-1})$ \\
		$V_{x, H-1} \gets \nabla_{x}\phi(\bar{\xx}_{H-1})$ \\
		$V_{xx, H-1} \gets \nabla_{xx}\phi(\bar{\xx}_{H-1})$ \\
		
		\For{$t \gets H-2$ \KwTo $0$}
		{
			Calculate $Q$ and derivatives at $t$ using \ref{eq:Q_and_derivs} \\
			Calculate $K^{fb}$ and $K^{ff}$ at $t$ \\
			Calculate $V$ and derivatives at $t$ using \ref{eq:bw_first} - \ref{eq:bw_last} \\
		}
		
		\For{$t \gets 1$ \KwTo $H-1$}
		{
			$\delta \xx \gets \bar{\xx}_{t} - \bar{\xx}_{t-1}$ \\
			$\delta \uu \gets K^{ff}_{t-1} + K^{fb}_{t-1}\delta \xx$ \\
			$\bar{\uu}_{t-1} \gets \bar{\uu}_{t-1} + \alpha \delta \uu$ \\
			$\bar{\xx}_t \gets \bar{\xx}_{t-1} + f(\bar{\xx}_{t-1}, \bar{\uu}_{t-1}) \Delta t$
		}
		
		return $\bar{\uu}_{0:H-2}$, $\bar{\xx}_{0:H-1}$, $K^{fb}_{0:H-2}$, $K^{ff}_{0:H-2}$
	}
\end{algorithm}

\subsection{Epistemic and Aleatoric Uncertainty}
As indicated in \cite{Kendall2017}, uncertainty in machine learning models can come from two sources: incomplete data and incomplete knowledge of the environment.  The former, called \emph{epistemic uncertainty} is the result of a data set that is either not large enough or does not cover enough of the ``search space''.  Such uncertainty could be explained away by access to unlimited data and can be reduced by the collection of additional data.  The latter source, called \emph{aleatoric uncertainty}, is the result of our inability to completely sense all aspects of the environment.  While such uncertainty could not be explained away with access to more data -- or even with access to unlimited data -- it could be explained away with access to unlimited sensing.

Note that in both cases, completely explaining away the uncertainty requires access to something we cannot possible obtain, i.e., either unlimited data or unlimited sensing capability.  It is therefore useful to consider using these sources of uncertainty in an engineering or machine learning context.  Epistemic uncertainty measures what our model doesn't know; given data, we should be able to train a model that can output this type of uncertainty.  Aleatoric uncertainty measures noise that is inherent in the environment, i.e., it measures what cannot be explained by data.  Again, given data, we should be able to train a model to output this type of uncertainty as well.  In principle, the total predictive uncertainty would be the combination of the two.  As shown in \cite{Kendall2017}, it is sometimes possible to use just one of these measures to develop a reasonable model; however, the more principled approach of combining these predictive uncertainty estimates often results in a superior model.

\subsection{Bayesian Approximation with Dropout}
Bayesian learning methodologies in and with neural networks have a long history \cite{Neal1995}.  Currently, at least two popular methods of Bayesian learning with neural networks appear frequently in literature.  The first, which essentially produces predictive distributions resulting from a distribution over neural network structure, is the Dropout approach introduced in \cite{Gal2016Bayesian}.  The second, in which predictive distributions are obtained due to probability distributions over network weights, is the Weight Uncertainty approach introduced in \cite{pmlr-v37-blundell15}.  The two approaches are similar in a number of ways, not least of which is the need to impose a variational approximating distribution over network weights and perform Monte Carlo sampling.  We discuss the dropout technique, which we elect to use in our work; the alternative approach requires doubling the number of parameters in the network, making it difficult to use for real-time MPC for vehicles with only onboard computation given very large convolutional neural networks.

Dropout has traditionally been used as a regularization method to prevent overfitting while training DNNs \cite{Hinton2012}.  It does so by effectively annihilating connections between neurons, resulting in a network that is much simpler, i.e., has fewer effective model parameters and is thus less prone to overfitting training data.  This is achieved by randomly sampling a binary \emph{dropout mask} for each layer $i$ from a Bernoulli distribution with probability of success $p_i$ and applying this mask to the weights of layer $i$.  For example, a simple two-layer feedforward network is represented by the equation

\begin{align}
	\mathbf{y} = \mathbf{W}_2 \sigma(\mathbf{W}_1 \mathbf{x} + \mathbf{b_1}),
\end{align}

where $\sigma$ is an component-wise activation function and the network has outputs $\mathbf{y} \in \mathbb{R}^K$, inputs $\mathbf{x} \in \mathbb{R}^D$, weight matrices $W_1 \in \mathbb{R}^{N \times D}$ and $W_2 \in \mathbb{R}^{K \times N}$, and bias $\mathbf{b}_1 \in \mathbb{R}^N$ (output bias has been neglected for clarity).  Binary dropout masks $\mathbf{z}_1 \in \{0, 1\}^D$ and $\mathbf{z}_2 \in \{0, 1\}^N$ can be sampled component-wise from Bernoulli distributions with success probability $p_1$ and $p_2$ for layers $1$ and $2$, respectively, and applied to the network resulting in a structurally simpler model with some connections removed.

\begin{align}
	\mathbf{\hat{y}} = \big[\mathbf{W}_2 diag(\mathbf{z}_2)\big] \sigma(\big[\mathbf{W}_1 diag(\mathbf{z}_1)\big] \mathbf{x} + \mathbf{b_1})
\end{align}

Dropout as a method of Bayesian approximation differs from the traditional approach in two respects.  First, as presented in \cite{Gal2016Bayesian}, dropout in this context is randomly applied at test time as well as during training.  This means that the predictive output for a single forward pass with a single input is the result of a network that is structurally simpler than the original.  Since the random nature of the dropout masks means that the same test-time input will result in a different output each time the input is passed through the network, several samples of the output corresponding to the same input result in a distribution that estimates epistemic uncertainty.  In addition, the dropout probability $p$ for each layer can be learned along with the network weights during training using the Concrete Dropout approach \cite{Gal2017Concrete}.

Aleatoric uncertainty is not learned via sampling, but instead learned by using the so-called \emph{heteroscedastic loss} function (described below) and splitting the output layer of the network (Fig. \ref{fig:DropoutVGG}) such that one channel is trained to output a predictive mean and the other the aleatoric uncertainty \cite{Kendall2017}.  By combining the uncertainty obtained via sampling with that obtained during training, an accurate representation of total model uncertainty is obtained.

\subsection{Batch Imitation Learning}
Let $\mathcal{S}, \mathcal{O}, \mathcal{U}$ be the state, observation, and admissable control spaces of the task under consideration, respectively.  Our goal is to determine a neural network policy $\pi: \mathcal{O} \rightarrow \mathcal{U}$ that is able to minimize the expected accumulated cost over a discrete finite time horizon $H$:

\begin{align}
	\min_{\pi} \mathbb{E}_{p_{\pi}} \Big[ \sum_{t=0}^{H-1} l(\xx_t, \uu_t) \Big],
\end{align}

where $\xx_t \in \mathcal{S}$, $\uu_t \in \mathcal{U}$, and running (immediate) cost $l$ for consistency with the notation introduced earlier.  $p_{\pi}$ is the joint distribution of $\xx_t$, $\uu_t$, and $\mathbf{o}_t \in \mathcal{O}$ for the policy $\pi$ for $t = 0, \ldots, H-1$.

This cost may be unavailable or impossible for a learned model to optimize directly.  Instead, imitation learning assumes that an expert capable of optimizing this cost is available.  The goal of imitation learning is to learn a policy that minimizes the deviation between the cost of the learned model and that incurred by the expert policy $\pi^*$

\begin{align}
	\pi_{NN} = \arg\min_\pi \mathbb{E}_{p_{\pi^*}} \Big[l(\xx_t, \uu_t) \Big],
\end{align}

where $\uu_t = \pi(\mathbf{o}_t)$ and $\pi_{NN}$ denotes the selected neural network policy.  In batch imitation learning, as used in this work, this is reduced to a supervised learning problem in which the model is trained to learn from data, consisting of observation-action or state-action pairs, generated by the expert.

End-to-end learning offers a number of advantages; our primary motivation in this work is that end-to-end methods avoid the need for burdensome manual engineering and unprincipled parameter tuning.

\section{Safe End-to-End Learning for Model Predictive Control of Autonomous Vehicles} \label{sec:safe_imlearn_control}
We wish exploit end-to-end learning to train a neural network policy to perform MPC for autonomous vehicle control. The trained model must be safe; in our framework, safety is introduced by returning control to an expert controller when the learned model is unable to produce a suitable output.  Such a framework is useful for long-term missions in which it is infeasible or impossible to continuously use an expert.  This could be the case, for example, when the power and computational requirements of an iterative optimal controller or the natural limitations of a human expert become a burden on the system.

As noted above, our probabilistic policy is a deep Bayesian neural network using the dropout technique.  We apply dropout to all weight-containing layers in our models.  At test time we perform Monte Carlo sampling of the output by repeatedly passing in the same input, thereby estimating epistemic uncertainty by calculation of the sample distribution.  The linear output layers of our models are split, with one channel trained to calculate aleatoric uncertainty.  In order to do so, we train the model with the heteroscedastic loss introduced in \cite{Kendall2017}

\begin{equation} \label{eq:loss function}
	Loss = \sum_i \frac{\norm{y_i - \mu}_2^2}{\sigma^2} + \log \sigma^2,
\end{equation}

where the network predicts $\mu, \sigma^2$ and the true target is $y$.  We recognize this as being essentially equivalent to the scalar Gaussian log likelihood of independent data samples.

Our expert controller in all tasks considered herein is a real-time MPC controller operating between 40Hz and 60Hz which applies the DDP trajectory optimization algorithm in a receding horizon fashion.  In all cases, the MPC-DDP expert has access to full state information, i.e., is assumed to have perfect knowledge of the state during the generation of training data.  The learned model may have only partial state information depending on the specific task, each of which is discussed in subsequent sections.

In keeping with the motivation to avoid manual selection of model parameters and hyperparameters, the dropout probability for each layer is trained along with network weights using the Concrete Dropout approach.  In contrast to competing methods, our framework also avoids hand-tuned specification of the uncertainty threshold above which the MPC-DDP expert regains control of the system.

\subsection{Reinforcement Learning of the Uncertainty Threshold}
High predictive uncertainty generated by the Bayesian network is indicative of a novel test-time input, i.e., data of a type very different from the distribution of the training data.  When the output uncertainty is high, the learned model is incapable of producing a useful control; we therefore return control of the system to the MPC-DDP controller.  The threshold above which the switch from learned model to expert controller is manually selected in earlier works; here, we propose selecting the threshold in a principled manner with reinforcement learning.

In imitation learning, an expert will generally be more successful and obtain a larger reward (or lower cost) than a learned model for the same task.  As a result, naive application of RL to learn the uncertainty threshold will result in a policy that immediately hands control to the expert, i.e., it will learn a threshold value of 0 so that the expert's action is always queried for maximum reward.  Our approach therefore encourages the use of the learned model for as long as possible.

The choice of RL algorithm is important; our choice is motivated by the observation that the switch from the Bayesian neural network policy to the MPC-DDP expert is best viewed as a rare discrete event.  The Cross-Entropy Method (CEM) \cite{DeBoer2005} applied to reinforcement learning is well-suited to such cases.  We use CEM to attempt optimization of the uncertainty threshold for switching from the neural network policy to the MPC-DDP expert as summarized in Algorithm \ref{alg:cem}.

Our results, discussed below, demonstrate that our framework converges to a threshold that allows the learned model to perform the task while switching back to the expert while still allowing the expert enough time to recover from or handle the unusual inputs.

\begin{algorithm}
	\caption{Cross-Entropy Method to optimize switching uncertainty threshold}
	\label{alg:cem}
	\SetKwInOut{Input}{Input}
	\SetKwInOut{Output}{Output}
	
	\Input{$x$ : initial state, \\
		$n_{RL}, n_{ro}$ : episodes \& rollouts per episode,\\
		$T$ : maximum time steps for task completion, \\
		$\mu, \sigma^2$ : initial distribution of the threshold, \\
		$r, \phi$ : immediate \& terminal reward functions, \\
		$\underline{b}, \bar{b}$ : min \& max number of candidates, \\
		$B$ : set of $\bar{b}$ best candidate solutions, \\
		$\pi_L, \pi_E$ : NN and MPC-DDP policies
	} 
	\Output{$\mu^*$ : optimized mean value of the threshold, \\
		$R_{sum}$ : sum of rewards \\
	}
	
	\For{$i \gets 1$ \KwTo $n_{RL}$}
	{
		Reset the environment and $x$ \\
		$success \gets 0$ \\
		Obtain $n_{ro}$ samples $X = [X(1), \ldots, X(n_{ro})] \sim \mathcal{N}(\mu, \sigma^2)$ \\
		\For{$j \gets 1$ \KwTo $n_{ro}$}
		{
			$R_j \gets 0$,~
			actor $\gets \pi_L$ \\
			Execute actions from $\pi_L$ until an adversarial input occurs at time step $t'$ \\
			\For{$t \gets t'$ \KwTo $T-1$}
			{
				Execute actor action \\
				actor $\gets \pi_E$ if uncertainty $ > X(j)$ \\
				$R_j \gets R_j + r(t)$\\
			}
			$R_j \gets R_j + \phi(T)$ \\
			Increment $success$ if $\phi(T) \neq 0$\\
			
			sort $X$ by descending $R_j$ \\
			$B \gets X[1:\bar{b}]$ \\
			$(\mu, \sigma^2) \gets (mean(B), var(B))$ \\
		}
		Repeat from Step 3 if $success < \underline{b}$ \\
		
		
		$R_{sum}(i) \gets \Sigma_{j=1}^{n_{rollout}} R_j$\\
	}
	return $\mu$, $R_{sum}$
\end{algorithm}

\section{Experiments} \label{sec:experiments}
We validate our framework for safe learning of model predictive control on two different tasks:
\begin{enumerate}
	\item Cart-pole swing-up with full and partial state observability
	\item Vision-based rally car driving with obstacle avoidance
\end{enumerate}

\begin{figure*}
	\centering
	\subfloat{{\includegraphics[width=0.75\textwidth]{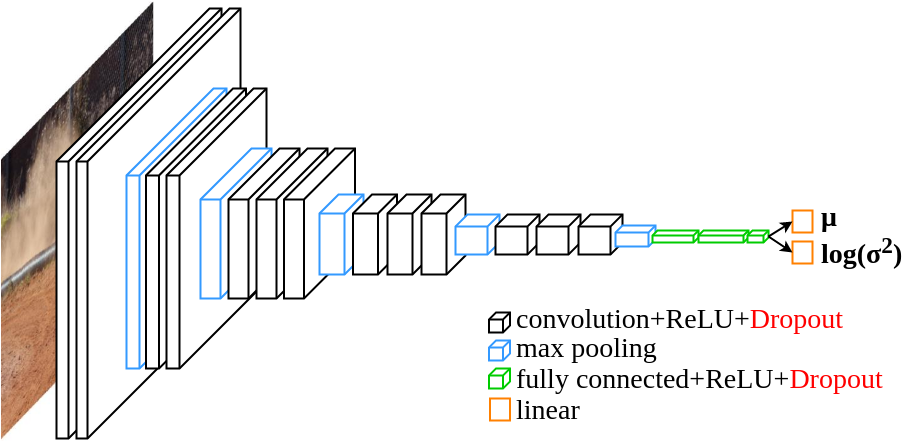}}}
	\caption{DropoutVGG Net}
	\label{fig:DropoutVGG}
\end{figure*}

For vision-based tasks, we introduce the DropoutVGG Network.  This Bayesian extension to VGG Net \cite{VGG} uses the dropout technique as described above.  In addition, we split the output layer and train the network to output aleatoric uncertainty as shown in Figure \ref{fig:DropoutVGG}.

For all tasks, we provide positive terminal reward in the CEM algorithm if the task was completed successfully by the Bayesian network model or by the MPC-DDP controller.

\begin{align}\label{eq:terminal_reward}
\phi_T =
	\begin{cases}
		T, & \mbox{if success} \\
		0, & \mbox{otherwise}
	\end{cases}
\end{align}

In order to encourage use of the learned model, we provide a positive immediate reward for every time step for which the learned model is used.

\begin{align}
	r_t =
	\begin{cases}
		1, & \mbox{if using learner} \\
		0, & \mbox{if using DDP}
	\end{cases}
\end{align}

\subsection{Cart-Pole Swing-Up}
For the cart-pole swing-up task, the MPC-DDP expert was provided a system dynamics model identical to that used in the popular OpenAI Gym simulator \cite{gym}.  This system consists of a weighted pole balanced on a cart.  The goal is to swing up and balance the pole only by applying a horizontal force to the cart.  The terminal reward is typical of that used in OpenAI Gym; that is, we provide the reward in \ref{eq:terminal_reward} if at the terminal time $T$:

\begin{equation}
	-2.4 < x_{T} < 2.4 \enspace \text{and} \enspace -15 \degree <\theta_{T} < 15\degree,
\end{equation}

where $x$ is the horizontal position of the cart and $\theta$ is the pole angle.

\begin{figure*}
	\centering
	\subfloat{{\includegraphics[width=0.5\textwidth]{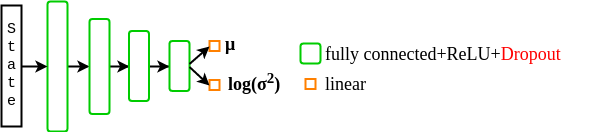}}}
	\caption{Feedforward Bayesian network with dropout}
	\label{fig:cartpolenn}
\end{figure*}

\subsubsection{Fully Observable Case}
In fully observable case, both the expert and the Bayesian neural network have perfect information regarding the complete system state $[x, \dot{x}, \theta, \dot{\theta}]^T$.  A convolutional model is not required as no vision-based control is performed; we instead use a simple feedforward Bayesian network (Fig. \ref{fig:cartpolenn}).

After training the neural network with the DDP expert, we allow the learned model to attempt the task.  The mass of the system is changed at an arbitrarily selected time to create a disturbance; the learner must revert control to the DDP controller before the angle of the pole falls outside the required limits and must do so in time for the DDP controller to recover.

\begin{figure*}
\centering
\subfloat[Total rewards per CEM episode]{{\includegraphics[width=0.4\textwidth]{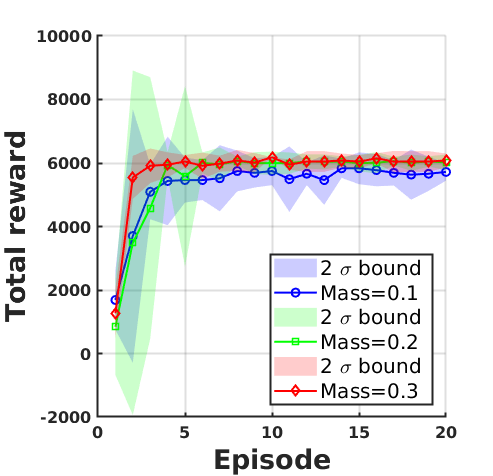} }}
\subfloat[Threshold per CEM episode]{{\includegraphics[width=0.4\textwidth]{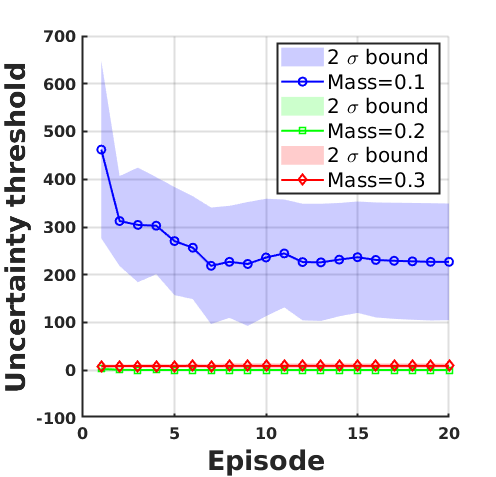} }}
\hspace{1mm}
\subfloat[Changed the cart mass to 0.1.]{{\includegraphics[width=0.33\textwidth]{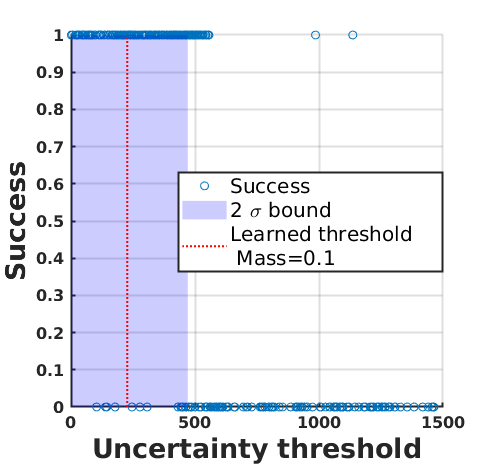} }}
\subfloat[Changed the cart mass to 0.2.]{{\includegraphics[width=0.33\textwidth]{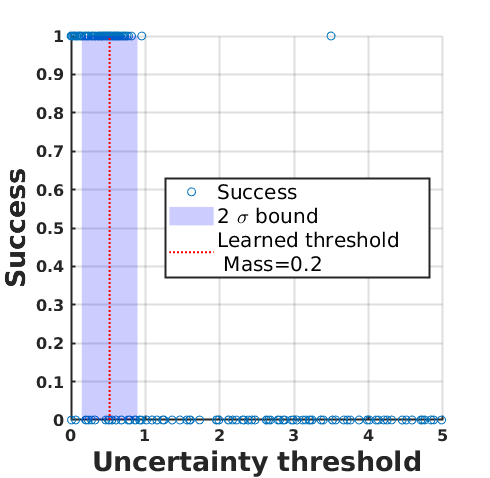} }}
\subfloat[Changed the cart mass to 0.3.]{{\includegraphics[width=0.33\textwidth]{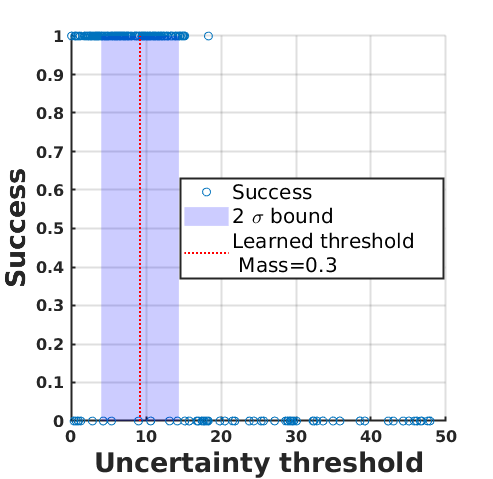} }}
\caption{Results of learning a switching threshold for the cart-pole swing-up task in the fully observable case for three different scenarios: a change of the cart mass to 0.1, 0.2, and 0.3 from 1.0, respectively. Bottom figures show success/failure trials per sample of the uncertainty threshold where we see that learned optimal thresholds are located between the success and failure margins.}
\label{fig:fo_cartpole}
\end{figure*}

\subsubsection{Partially Observable Case}
In the partially observable case, the neural network can estimate $x$ and $\theta$ but velocities $\dot{x}$ and $\dot{\theta}$ are not directly accessible.  DropoutVGG is trained with inputs consisting of (third-person view) gray-scale images of the cart-pole system and observations of the expert control; the learned model must use image information to infer the velocity components of the state in order to produce a usable control signal.

\begin{figure*}
\centering
\subfloat[Training data]{{\includegraphics[width=0.25\textwidth]{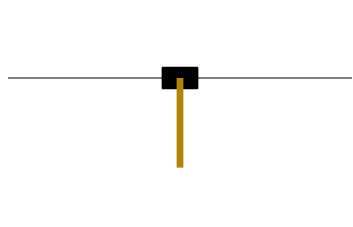} }}
\hspace{5mm}
\subfloat[Test with adversarial inputs]{{\includegraphics[width=0.75\textwidth]{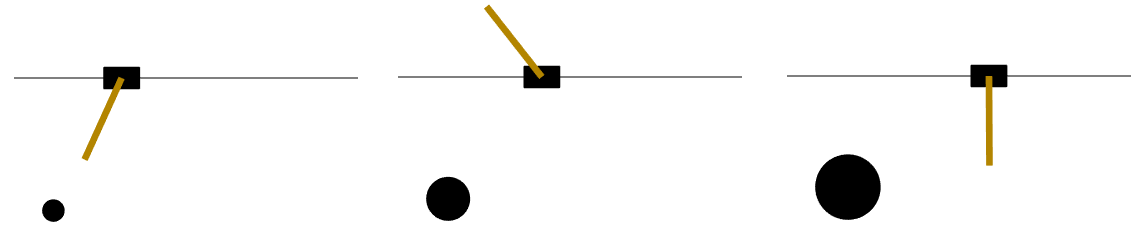} }}
\caption{Partially observable case of cart pole system.}
\end{figure*}

The goal and limits of the task are as the same as in the fully observable case.  In this case, however, a disturbance is created when the input image is randomly corrupted with the addition of a new object that was not present in the training data.

\begin{figure*}
\centering
\subfloat[Total rewards per CEM episode]{{\includegraphics[width=0.4\textwidth]{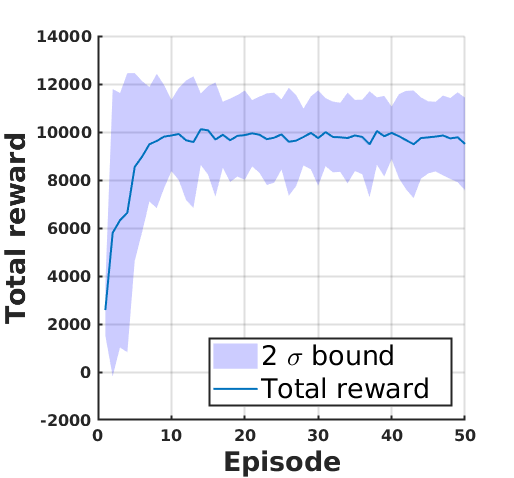} }}
\subfloat[Threshold per CEM episode]{{\includegraphics[width=0.4\textwidth]{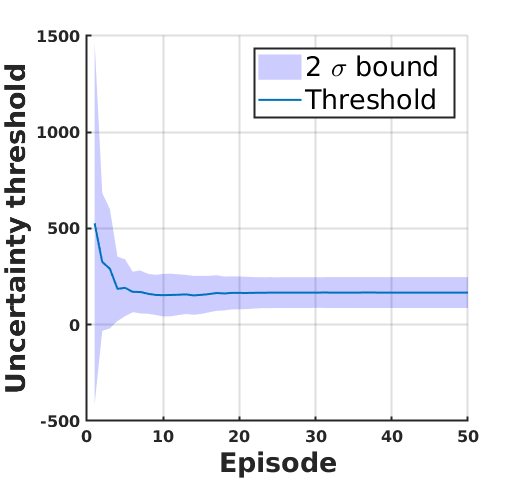} }}
\hspace{1mm}
\subfloat[Total reward per uncertainty threshold]{{\includegraphics[width=0.4\textwidth]{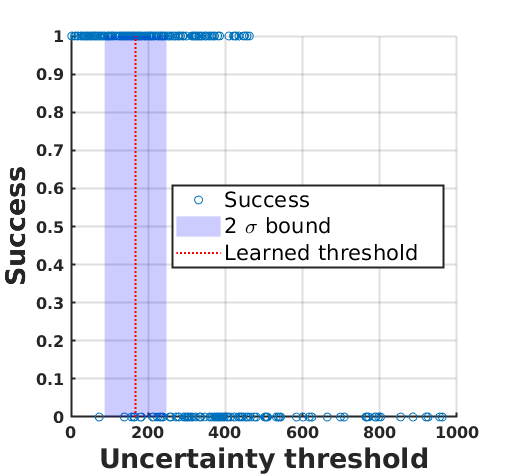} }}
\subfloat[Success per uncertainty threshold]{{\includegraphics[width=0.4\textwidth]{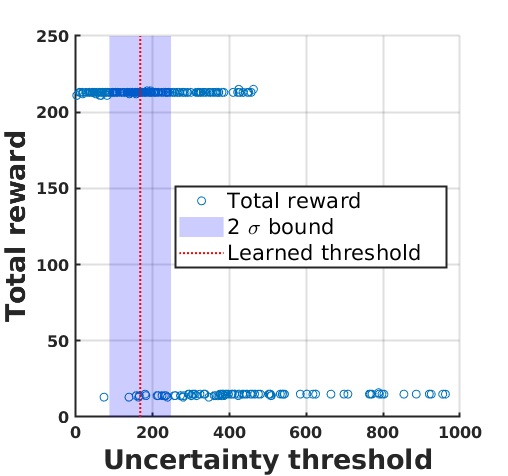} }}
\caption{Results of learning a switching threshold for a cart pole swingup task in partially observable case.  Interpretation is similar to Fig. \ref{fig:fo_cartpole}.}
\end{figure*}

Interestingly, we found that the predictive uncertainty of the neural network grows when larger objects corrupt the input image.  Tests with color images also showed extreme sensitivity to the color of the new object; we hope to investigate this further, but for our current results the image was converted to grayscale before being passed to the neural network.

\subsection{Autonomous Navigation and Obstacle Avoidance}
\begin{figure*}
\centering
\subfloat[AutoRally simulation]{{\includegraphics[width=0.5\textwidth]{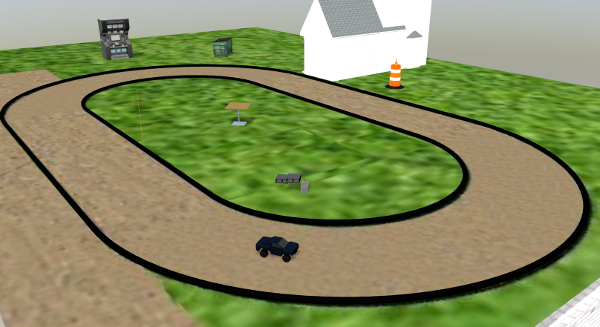} }}
\hspace{1mm}
\subfloat[Onboard view from the simulated vehicle]{{\includegraphics[width=0.45\textwidth]{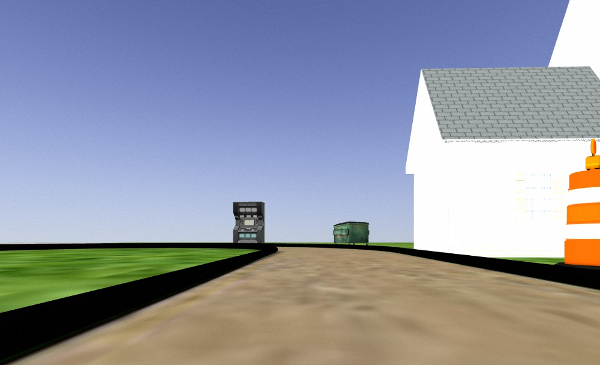} }}
\caption{AutoRally simulation environment}
\label{fig:gazebo}
\end{figure*}

Autonomous driving is a more realistic and significantly more complex task than cart-pole swing-up.  Here, we demonstrate the success of our framework on the autonomous navigation and obstacle avoidance task using the simulator provided by the AutoRally project \cite{AutoRallyGithub}.  This software \ref{fig:gazebo} simulates driving a 1/5th scale autonomous vehicle on a track and is capable of providing simulated color images from onboard cameras.  We again use DDP as the expert controller; in this experiment, the DDP controller uses a data-driven adaptive dynamics model based on Sparse Spectrum Gaussian Processes (SSGP) \cite{Pan-NIPS-16}.

DropoutVGG is trained with observations consisting of $80\times160$ pixel RGB images from the vehicle's left camera as well as the corresponding steering angle produced by the MPC-DDP expert as it navigates around the track at a speed of 5 m/s.  The goal is to train the network to learn to steer the vehicle; we do not learn the throttle (acceleration) in this experiment.  The learned model is not provided with any direct information about the vehicle state; it may only access the camera image.  As usual, no disturbances or unusual inputs are given to the neural network in the training phase.

In the absence of unusual test-time inputs, the Bayesian network easily learns to steer the vehicle based only on vision.  However, when a previously unseen obstacle comes into view (Fig. \ref{fig:uncertainty increase}) the predictive uncertainty increases significantly, indicating potential failure of the learned model.  Our method successfully trains the learned model to pass control back to the DDP expert in time to avoid the obstacle.

\begin{figure*}
\centering
\subfloat[Total rewards per CEM episode]{{\includegraphics[width=0.4\textwidth]{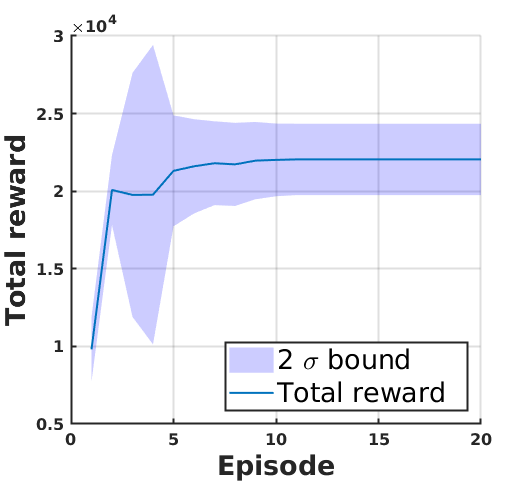} }}
\subfloat[Threshold per CEM episode]{{\includegraphics[width=0.4\textwidth]{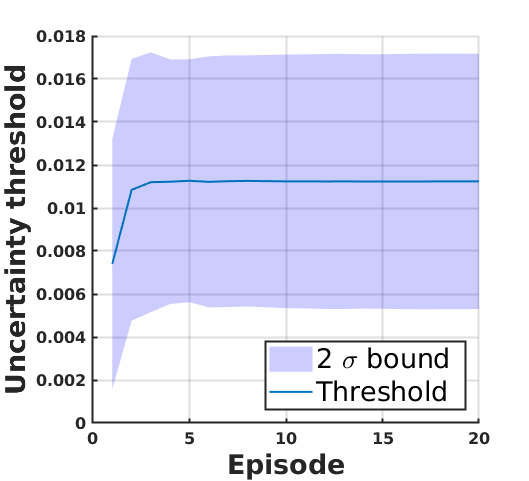} }}
\hspace{1mm}
\subfloat[Total reward per uncertainty threshold]{{\includegraphics[width=0.4\textwidth]{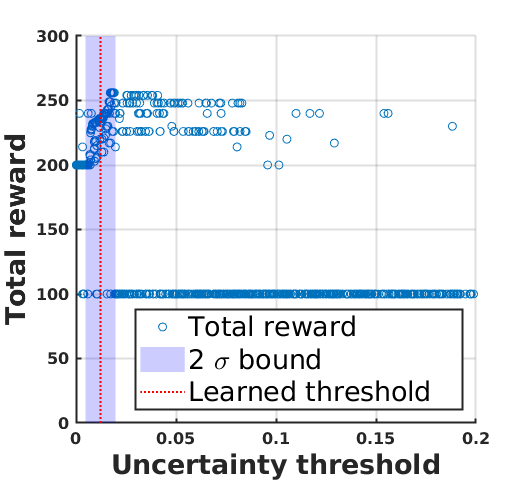} }}
\subfloat[Success per uncertainty threshold]{{\includegraphics[width=0.4\textwidth]{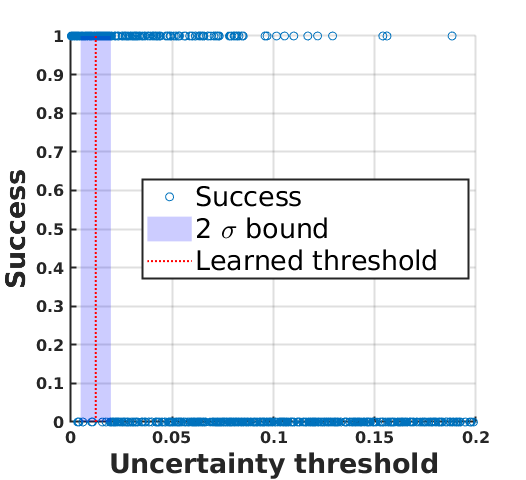} }}
\caption{Results of learning a switching threshold for an autonomous navigation and obstacle avoidance task in partially observable case, only using images.}
\end{figure*}

\section{Future Work} \label{sec:discussion}
As demonstrated above, the dropout approach for Bayesian approximation is a good fit for safe end-to-end learning for control.  However, sampling the network enough times to obtain accurate epistemic uncertainty estimates is time-consuming if done sequentially.  It may be parallelized but this is difficult for networks with a large number of weights.  We plan to investigate schemes for efficient sampling through Bayesian networks in a future work.
On a more theoretical note, we found during our experiments with the AutoRally simulator that the predictive uncertainty increases far later than would be ideal; a promising avenue of research is to investigate recurrent or generative models capable of propagating predictive uncertainty into the future in order to obtain better behavior, especially in high-speed tasks.
Another future direction is the extension of the current work to online imitation learning.  As noted above, our current framework uses batch imitation learning, essentially reducing the problem to a simple supervised learning problem.  However, as shown in \cite{DAgger}, the DAgger algorithm and its variants can significantly reduce the regret bounds relative to this type of algorithm.  We continue to investigate the best way to combine our framework -- especially the RL component -- with the DAgger methodology.

\section{Conclusion} \label{sec:conclusion}
We have presented herein a fully end-to-end framework for safe learning of model predictive control.  Our approach successfully trains a Bayesian neural network to perform complex vision-based MPC with access to a dataset containing image observations and corresponding controls from an expert based on differential dynamic programming.  This approach allows the learned model to effectively know when it will be unable to safely complete the task and return control to an expert capable of doing so.  Our framework requires no manual parameter tuning; instead we propose the use of a reinforcement learning method to learn uncertainty thresholds in a principled manner.  We validate our method on cart-pole swing-up and autonomous driving tasks.

\bibliographystyle{IEEEtran}
\bibliography{references}

\begin{thebibliography}{10}
\providecommand{\url}[1]{#1}
\csname url@samestyle\endcsname
\providecommand{\newblock}{\relax}
\providecommand{\bibinfo}[2]{#2}
\providecommand{\BIBentrySTDinterwordspacing}{\spaceskip=0pt\relax}
\providecommand{\BIBentryALTinterwordstretchfactor}{4}
\providecommand{\BIBentryALTinterwordspacing}{\spaceskip=\fontdimen2\font plus
\BIBentryALTinterwordstretchfactor\fontdimen3\font minus
  \fontdimen4\font\relax}
\providecommand{\BIBforeignlanguage}[2]{{%
\expandafter\ifx\csname l@#1\endcsname\relax
\typeout{** WARNING: IEEEtran.bst: No hyphenation pattern has been}%
\typeout{** loaded for the language `#1'. Using the pattern for}%
\typeout{** the default language instead.}%
\else
\language=\csname l@#1\endcsname
\fi
#2}}
\providecommand{\BIBdecl}{\relax}
\BIBdecl

\bibitem{Bishop2006}
C.~M. Bishop, \emph{{Pattern recognition and machine learning}}.\hskip 1em plus
  0.5em minus 0.4em\relax Springer, 2006.

\bibitem{DAgger}
\BIBentryALTinterwordspacing
S.~Ross, G.~J. Gordon, and J.~A. Bagnell, ``A reduction of imitation learning
  and structured prediction to no-regret online learning,'' in
  \emph{Proceedings of the 14th International Conference on Artificial
  Intelligence and Statistics}, ser. JMLR, vol.~15, Fort Lauderdale, FL, USA,
  2011. [Online]. Available:
  \url{http://proceedings.mlr.press/v15/ross11a/ross11a.pdf}
\BIBentrySTDinterwordspacing

\bibitem{Pan-NIPSWS-17}
\BIBentryALTinterwordspacing
Y.~Pan, C.-A. Cheng, K.~Saigol, K.~Lee, X.~Yan, E.~Theodorou, and B.~Boots.,
  ``Learning deep neural network control policies for agile off-road autonomous
  driving,'' in \emph{The NIPS Deep Rienforcement Learning Symposium}, 2017.
  [Online]. Available:
  \url{https://www.cc.gatech.edu/~bboots3/files/nips17drl.pdf}
\BIBentrySTDinterwordspacing

\bibitem{Pereira2018}
\BIBentryALTinterwordspacing
M.~Pereira, D.~D. Fan, G.~N. An, and E.~Theodorou, ``{MPC-Inspired Neural
  Network Policies for Sequential Decision Making},'' feb 2018. [Online].
  Available: \url{http://arxiv.org/abs/1802.05803}
\BIBentrySTDinterwordspacing

\bibitem{Bojarski2017}
\BIBentryALTinterwordspacing
M.~Bojarski, P.~Yeres, A.~Choromanska, K.~Choromanski, B.~Firner, L.~Jackel,
  and U.~Muller, ``{Explaining How a Deep Neural Network Trained with
  End-to-End Learning Steers a Car},'' apr 2017. [Online]. Available:
  \url{http://arxiv.org/abs/1704.07911}
\BIBentrySTDinterwordspacing

\bibitem{Strickland-17}
\BIBentryALTinterwordspacing
M.~Strickland, G.~E. Fainekos, and H.~B. Amor, ``Deep predictive models for
  collision risk assessment in autonomous driving,'' \emph{CoRR}, vol.
  abs/1711.10453, 2017. [Online]. Available:
  \url{http://arxiv.org/abs/1711.10453}
\BIBentrySTDinterwordspacing

\bibitem{DropoutDAgger}
\BIBentryALTinterwordspacing
K.~Menda, K.~Driggs-Campbell, and M.~J.~Kochenderfer, ``Dropoutdagger: A
  bayesian approach to safe imitation learning,'' 09 2017. [Online]. Available:
  \url{https://arxiv.org/abs/1709.06166}
\BIBentrySTDinterwordspacing

\bibitem{Gal2016Bayesian}
\BIBentryALTinterwordspacing
Y.~Gal and Z.~Ghahramani, ``Dropout as a bayesian approximation: Representing
  model uncertainty in deep learning,'' in \emph{Proceedings of The 33rd
  International Conference on Machine Learning}, ser. Proceedings of Machine
  Learning Research, vol.~48.\hskip 1em plus 0.5em minus 0.4em\relax New York,
  New York, USA: PMLR, 20--22 Jun 2016, pp. 1050--1059. [Online]. Available:
  \url{http://proceedings.mlr.press/v48/gal16.html}
\BIBentrySTDinterwordspacing

\bibitem{VGG}
\BIBentryALTinterwordspacing
K.~Simonyan and A.~Zisserman, ``Very deep convolutional networks for
  large-scale image recognition,'' \emph{CoRR}, vol. abs/1409.1556, 2014.
  [Online]. Available: \url{http://arxiv.org/abs/1409.1556}
\BIBentrySTDinterwordspacing

\bibitem{Gal2017Concrete}
\BIBentryALTinterwordspacing
Y.~Gal, J.~Hron, and A.~Kendall, ``Concrete dropout,'' in \emph{Advances in
  Neural Information Processing Systems 30 (NIPS)}, 2017. [Online]. Available:
  \url{https://arxiv.org/abs/1705.07832}
\BIBentrySTDinterwordspacing

\bibitem{MPCDDP}
\BIBentryALTinterwordspacing
Y.~Tassa, T.~Erez, and W.~D. Smart, ``Receding horizon differential dynamic
  programming,'' in \emph{Advances in Neural Information Processing Systems
  20}, J.~C. Platt, D.~Koller, Y.~Singer, and S.~T. Roweis, Eds.\hskip 1em plus
  0.5em minus 0.4em\relax Curran Associates, Inc., 2008, pp. 1465--1472.
  [Online]. Available:
  \url{http://papers.nips.cc/paper/3297-receding-horizon-differential-dynamic-programming.pdf}
\BIBentrySTDinterwordspacing

\bibitem{Kendall2017}
\BIBentryALTinterwordspacing
A.~Kendall and Y.~Gal, ``{What Uncertainties Do We Need in Bayesian Deep
  Learning for Computer Vision?}'' mar 2017. [Online]. Available:
  \url{http://arxiv.org/abs/1703.04977}
\BIBentrySTDinterwordspacing

\bibitem{Neal1995}
R.~M. Neal, ``Bayesian learning for neural networks.''\hskip 1em plus 0.5em
  minus 0.4em\relax PhD Thesis, University of Toronto, 1995.

\bibitem{pmlr-v37-blundell15}
\BIBentryALTinterwordspacing
C.~Blundell, J.~Cornebise, K.~Kavukcuoglu, and D.~Wierstra, ``Weight
  uncertainty in neural network,'' in \emph{Proceedings of the 32nd
  International Conference on Machine Learning}, ser. Proceedings of Machine
  Learning Research, F.~Bach and D.~Blei, Eds., vol.~37.\hskip 1em plus 0.5em
  minus 0.4em\relax Lille, France: PMLR, 07--09 Jul 2015, pp. 1613--1622.
  [Online]. Available: \url{http://proceedings.mlr.press/v37/blundell15.html}
\BIBentrySTDinterwordspacing

\bibitem{Hinton2012}
\BIBentryALTinterwordspacing
G.~E. Hinton, N.~Srivastava, A.~Krizhevsky, I.~Sutskever, and R.~R.
  Salakhutdinov, ``{Improving neural networks by preventing co-adaptation of
  feature detectors},'' jul 2012. [Online]. Available:
  \url{http://arxiv.org/abs/1207.0580}
\BIBentrySTDinterwordspacing

\bibitem{DeBoer2005}
\BIBentryALTinterwordspacing
P.-T. de~Boer, D.~P. Kroese, S.~Mannor, and R.~Y. Rubinstein, ``{A Tutorial on
  the Cross-Entropy Method},'' \emph{Annals of Operations Research}, vol. 134,
  no.~1, pp. 19--67, feb 2005. [Online]. Available:
  \url{http://link.springer.com/10.1007/s10479-005-5724-z}
\BIBentrySTDinterwordspacing

\bibitem{gym}
\BIBentryALTinterwordspacing
G.~Brockman, V.~Cheung, L.~Pettersson, J.~Schneider, J.~Schulman, J.~Tang, and
  W.~Zaremba, ``Openai gym,'' 2016. [Online]. Available:
  \url{https://arxiv.org/abs/1606.01540}
\BIBentrySTDinterwordspacing

\bibitem{AutoRallyGithub}
``Autorally,'' \url{http://autorally.github.io/}, 2016.

\bibitem{Pan-NIPS-16}
\BIBentryALTinterwordspacing
Y.~Pan, X.~Yan, E.~Theodorou, and B.~Boots., ``Adaptive probabilistic
  trajectory optimization via efficient approximate inference,'' in \emph{29th
  Conference on Neural Information Processing Systems (NIPS)}, Barcelona,
  Spain, 2016. [Online]. Available: \url{https://arxiv.org/pdf/1608.06235.pdf}
\BIBentrySTDinterwordspacing

\end{thebibliography}

\end{document}